%% file: main_tex.tex
\documentclass[conference]{IEEEtran}
\IEEEoverridecommandlockouts
\usepackage{amssymb}  
\usepackage{pifont}   

\usepackage{cite}
\usepackage{amsmath,amssymb,amsfonts}
\usepackage{algorithmic}
\usepackage{graphicx}
\usepackage{booktabs}
\usepackage{multirow}
\usepackage[linesnumbered,ruled,vlined]{algorithm2e}
\usepackage{textcomp}
\usepackage{listings}
\usepackage{xcolor}
\usepackage{multirow} 
\usepackage{array} 

\def\BibTeX{{\rm B\kern-.05em{\sc i\kern-.025em b}\kern-.08em
    T\kern-.1667em\lower.7ex\hbox{E}\kern-.125emX}}
\begin{document}

\title{Learning to unfold cloth: Scaling up world models to deformable object manipulation}

\author{\IEEEauthorblockN{1\textsuperscript{st} Jack Rome}
\IEEEauthorblockA{
\textit{University of Edinburgh}\\
j.a.rome@ed.ac.uk}
\and
\IEEEauthorblockN{2\textsuperscript{nd} Stephen James}
\textit{Imperial College London}\\
\and
\IEEEauthorblockN{3\textsuperscript{rd} Subramanian Ramamoorthy}
\textit{University of Edinburgh}\\

\thanks{This work was supported by a postgraduate studentship from Dyson Technology Ltd. For the purpose of open access, the author(s) has applied a Creative Commons Attribution (CC BY) license to any Accepted Manuscript version arising.}

}

\maketitle

\input{sections/abstract_text}


\input{sections/introduction}
\input{sections/background}
\input{sections/methodology}
\input{sections/results}
\input{sections/conclusion}
\input{sections/future_work}

\medskip
\bibliographystyle{IEEEtran}
\bibliography{bibliography}

\end{document}

%% file: sections/abstract_text.tex
\begin{abstract}

Learning to manipulate cloth is both a paradigmatic problem for robotic research and a problem of immediate relevance to a variety of applications ranging from assistive care to the service industry. The complex physics of the deformable object makes this problem of cloth manipulation nontrivial. In order to create a general manipulation strategy that addresses a variety of shapes, sizes, fold and wrinkle patterns, in addition to the usual problems of appearance variations, it becomes important to carefully consider model structure and their implications for generalisation performance. In this paper, we present an approach to \textit{in-air} cloth manipulation that uses a variation of a recently proposed reinforcement learning architecture, DreamerV2. Our implementation modifies this architecture to utilise surface normals input, in addition to modiying the replay buffer and data augmentation procedures. Taken together these modifications represent an enhancement to the world model used by the robot, addressing the physical complexity of the object being manipulated by the robot. We present evaluations both in simulation and in a zero-shot deployment of the trained policies in a physical robot setup -- performing in-air unfolding of a variety of different cloth types -- demonstrating the generalisation benefits of our proposed architecture. 
\end{abstract}

%% file: sections/introduction.tex
\section{introduction}
Cloth manipulation is a seemingly simple task that features throughout our daily lives, from dressing in the morning to handling drapery and bedding. However, these tasks are extremely challenging for robotic systems, with few autonomous robotic systems being capable of performing such tasks `in the wild'. Within the robotics literature, cloth-related tasks have been approached with two categories of motivating application: assistive care tasks, such as dressing and undressing, and organizational tasks, like loading a washing machine, folding, or hanging up clothes \cite{puthuveetil2022bodies}, \cite{erickson2020assistivegym}, \cite{clegg2020learning}.

Manipulating cloth objects is particularly challenging for robots due to the complex deformations that occur during handling, which make it difficult to accurately track the shape and {\textit{state}} of the cloth. Frequently, the desired grasp points are obscured by overlapping fabric, and the surface on which the cloth rests may act as an obstacle, preventing direct access to these areas. Traditional methods for cloth unfolding involve multiple manipulations to expose the target area for grasping \cite{wu2019learning}. However, an alternative approach is to lift the cloth into the air, allowing the fabric to drape and making the target area more accessible for manipulation. The issue now is that the task has become $3-$dimensional, in contrast to the traditional on-table methods, adding further complexity to the manipulation task. In particular, the autonomoous robot must now reason about the complete dynamics of the $3-$dimensional object, as opposed to the quasi-static configuration changes of a nearly-planar object. 

\begin{figure}[tp] 
    \centering
    \includegraphics[width=1.1\linewidth]{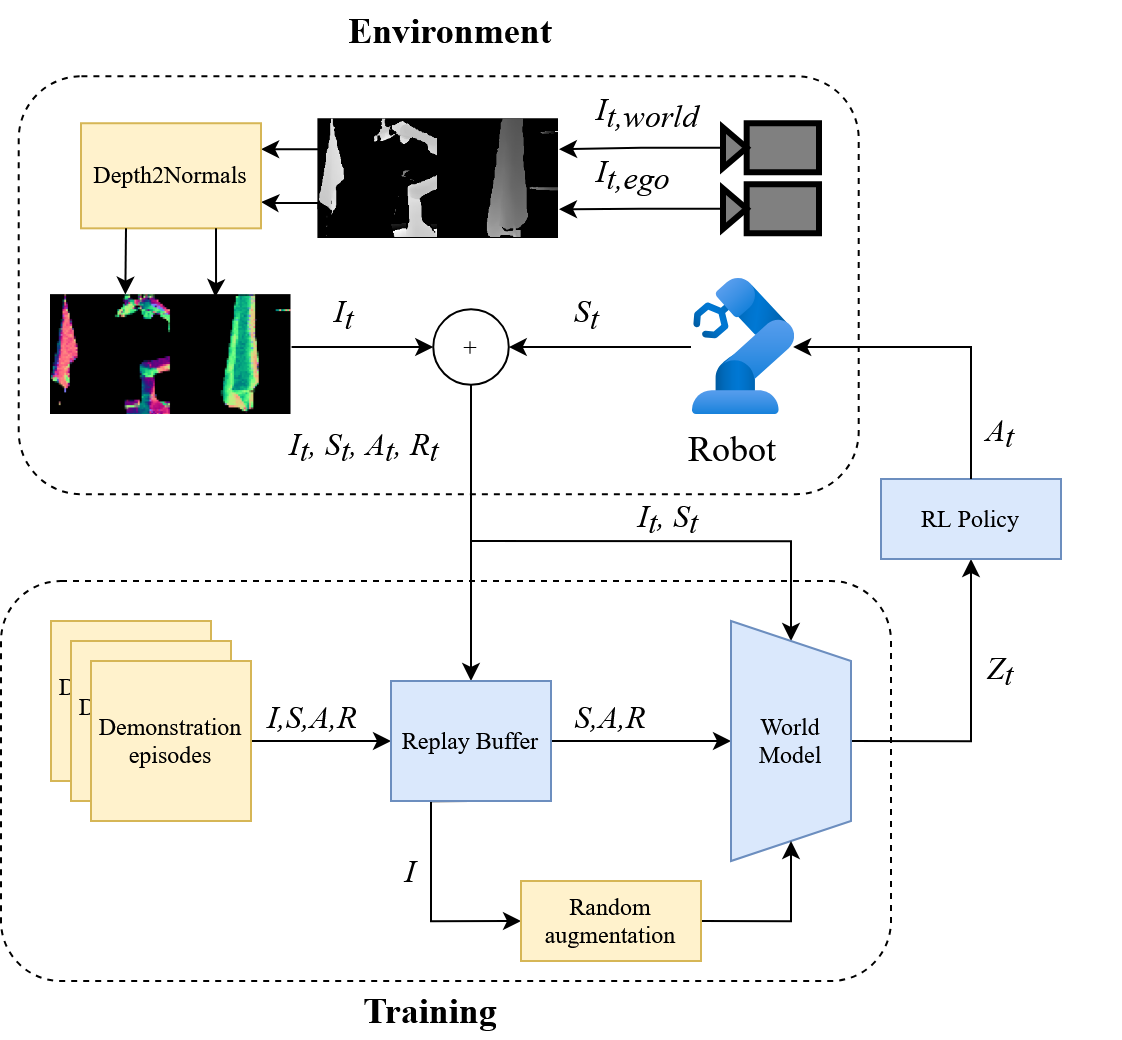}
    \caption{System diagram of learning architecture with our additions in yellow. Where image state action reward trajectories are denoted as I,S,A,R and latent space, Z. Depth images are  converted into surface normal images for image state, demonstration episodes are loaded into the replay buffer on start and batch random augmentation occurs while sampling images for training the world model}
    \label{fig:unfold demo}
\end{figure}

In robotic manipulation research, the case of rigid object models is well studied and is to an extent simpler because the object shape remains constant during interaction. The primary challenge with rigid bodies lies in determining optimal grasp points to secure the object and to avoid slippage during manipulation \cite{bicchi1995closure}. In contrast, deformable objects like cloth present a far greater challenge due to the complexity of the dynamics, with changes in configuration taking place at a much higher dimensionality. The cloth shape can change quickly and significantly during handling, requiring highly accurate environmental models and/or contollers robust enough to quickly adapt to modeling inaccuracies. However, such models often come with high computational costs, making them impractical for real-time use. Data-driven methods offer an alternative, as they can quickly find and execute optimal solutions for specific tasks. However, state-of-the-art machine learning methods in applied to this domain still struggle with fine-grained manipulation tasks, such as with deformable objects. One reason for this is the difficulty in representing and estimating {\textit{state}} and {\textit{dynamics}} of the environment and object in a way that allows for efficient calculation of optimal actions.

\input{sections/comparison_table}

Reinforcement learning has made significant advancements in recent years, demonstrating the ability to tackle increasingly complex robot manipulation tasks. One contributing factor has been the increasing sophistication of world models and their effective use in policy learning architectures. However, there is difficulty in using these methods to complete manipulation tasks on a diverse set of garments, such as in the home-setting, while also maintaining its ability to perform in realtime. Often this comes as a tradeoff. In this work, we utilize a state-of-the-art reinforcement learning model within our system for in-air unfolding of cloth objects. Building on the Dreamerv2 model \cite{hafner2020mastering}, we enhance it by replacing RGB input with calculated surface-normals which allow us to use a richer depth based model of the cloth in our policy calculations, emphasizing the geometry of the cloth while filtering out irrelevant visual appearance variations. Additionally, we modify the replay buffer to enhance the robustness of the world models and improve the system’s ability to generalize across cloth objects. The focus of our work is on fast and robust methods for in-air cloth unfolding and zero-shot sim-to-real deployment. 

%% file: sections/comparison_table.tex
\newcommand{\cmark}{\checkmark}
\newcommand{\xmark}{\ding{55}}

\begin{table*}[ht]
\centering
\begin{tabular}{|l|l|l|l|l|l|l|}
\hline
\textbf{Approach} & \textbf{Paper} & \textbf{Success Rate} & \textbf{Execution Time} & \textbf{No. Garments} & \textbf{No. Garment Types} & \textbf{Ablation Handling} \\
\hline
MB       & \cite{bender2013adaptive}                       & \xmark Low        & \xmark Slow          & \xmark Few        & \xmark Few            & \xmark No \\
MB       & \cite{wu2019learning}              & \cmark High       & \xmark Slow          & \cmark Several    & \xmark Few            & \xmark No \\
MB       &  \cite{li2018model}                            & \cmark Medium     & \xmark Slow          & \xmark Few        & \xmark Few            & \xmark No \\
MF        & \cite{matas2018sim}                  & \cmark Medium     & \cmark Real-time     & \cmark Several    & \cmark Several        & \xmark Limited \\
MF        & \cite{zhang2022learning}                  & \cmark Medium     & \cmark Real-time     & \cmark Several    & \cmark Moderate       & \xmark Limited \\
MF        & \cite{hietala2022learning}                       & \cmark High       & \cmark Real-time     & \cmark Moderate   & \cmark Several        & \cmark Yes \\
H   & \cite{gabas2021dual}                          & \cmark Medium     & \cmark Real-time     & \cmark Moderate   & \xmark Few            & \xmark No \\
H   &\cite{sunil2023visuotactile}                   & \cmark Medium     & \cmark Real-time     & \cmark Several    & \xmark Few            & \xmark No \\
Hybrid            & \cite{arnold2019fast}                                       & \cmark High       & \cmark Fast          & \cmark Several    & \cmark Moderate       & \cmark Partial \\
Hybrid            & \cite{ha2022flingbot}                                          & \cmark High       & \cmark Real-time     & \cmark Several    & \cmark Moderate       & \cmark Partial \\
\hline
\end{tabular}
\caption{Comparison of cloth manipulation methods across key metrics. Descriptive terms indicate approximate scale and generality of evaluation. Where MB, MF, H are Model-Free, Model-based, and heuristic based approaches}
\label{tab:cloth-comparison}
\end{table*}

%% file: sections/background.tex
\section{Related work}
A common starting point in the literature on cloth manipulation is to first process the state of the cloth to encode it within a latent space representation, which is then used as input to a network for semantic state estimation or policy computation. For example, in \cite{arnold2019fast}, the simulated environment provides a voxelized representation of the cloth, exaggerating the height to better utilize the voxel grid. This topological representation involves a substantial amount of data, specifically $32 \times 32 \times 16$, totaling over $16,000$ binary values. However, the researchers address this by employing an encoder-decoder network to reduce the state input to just $256$ floating-point values, making it a more manageable input for the "manipulation network" section of their pipeline. 

Similarly, raw image data can be too voluminous for direct network input without pre-processing, which is why image auto-encoders are often used. These auto-encoders learn to encode images into a lower-dimensional state, effectively reducing the data load while preserving essential features. This lower-dimensional state can be combined with other state information or sensor data, creating a multi-modal state space that provides richer environmental context for the system. For instance, combining vision with a robot's proprioception, as explored in \cite{lee2020making}, enhances the system's understanding and response to the environment.

Our objective is to achieve in-air unfolding of a cloth object, a critical step in tasks such as hanging laundry or folding clothes. This process is essential because a secure grasp on two opposite ends of the cloth is necessary to proceed with subsequent stages of a sequence of manipulation tasks. For instance, in the case of unfolding, it has been demonstrated in \cite{ha2022flingbot} that a well-executed dynamic fling can achieve an effective unfold, provided that the cloth is grasped properly at the outset. However, obtaining an optimal initial grasp is often challenging due to the self-occlusions that occur when the cloth is in a crumpled state, which can obscure the desired grasping areas, such as the corners. Instead of searching for both corners simultaneously, a more effective approach is to locate a viable grasp point and lift the cloth into the air. This action exposes the remaining corners, allowing a second manipulator to grasp them as they dangle freely.

In this work, we start from the situation wherein a successful grasp and lift has been accomplished, with the cloth dangling from one of its corners, and our task is to grasp another part of the cloth in order to perform an unfold. While the initial grasp and lift are critical and challenging, the specific focus of the work in this paper is on the subsequent unfolding process. Traditionally, this task is achieved using $3D$ sensors to reconstruct or process a full $3D$ representation of the cloth shape. This typically requires a series of depth sensors or careful rotation of the cloth to capture the $3D$ sensor readings from different angles without disrupting its shape \cite{li2018model},\cite{maitin2010cloth}. Although this method provides highly accurate state information, it comes with significant drawbacks -- including lengthy execution times and reliance on an accurate cloth model. With this process, it can take up to a minute to gather data on the cloth state, and any disruption during manipulation renders this model obsolete, necessitating a restart.

To address these limitations of current cloth manipulation techniques, we propose a method that can grasp and unfold cloth with high efficiency and speed. We develop a reinforcement learning approach to compute closed-loop policies, including continual reaction to changes to expected and unanticipated changes in the cloth’s state, in real-time \cite{jangir2020dynamic}. This adaptability and reactivity are expected to significantly improve performance over traditional $3D$ mapping-based methods, particularly when the task requires handling cloth in a dynamic setting.

Our approach significantly reduces execution time while maintaining increased flexibility. However, we recognize that implementing RL in this domain presents significant challenges. For example, this method requires extensive training data to learn the complex dynamics of in-air cloth manipulation, and designing effective reward functions will be crucial to guiding the RL model’s learning process. To mitigate these issues, we utilize a custom-built cloth simulator to effectively train our models with realistic cloth dynamics, including domain randomisation, before deploying in the physical system. 

We evaluate and demonstrate our method on metrics including execution speed, accuracy in cloth manipulation, and the system’s ability to adapt to various natural perturbations. Comparing the performance of our method against baselines demonstrates the practical advantages of our approach.

%% file: sections/methodology.tex
\section{Methodology}
\subsection{States, Actions, and Reward Function}
The simulated environment is developed using the Unity Game Engine. Data exchange between Unity’s C\# and Python is managed by a framework called RFUniverse \cite{Unity-Technologies_2022} \cite{fu2022rfuniverse}, which wraps the environment in a standard Gym interface, thereby making it accessible to the agent. Cloth dynamics are simulated via ObiCloth  \cite{Studio}, a Unity plugin. In this setup, one corner of the cloth is pinned so that it dangles in front of a Franka robot. At each timestep, the agent controls the robot’s end-effector displacement in the XYZ directions and toggles the gripper’s open/close state, where actions are scaled as A=[-1,1]*4. The agent’s task is to unfold the cloth by moving towards an adjacent corner of the pinned vertex, grasping the cloth, and pulling it outward. The main challenge is ensuring that the robot grasps a flat section of the cloth rather than a crumpled portion.

At every timestep, the agent receives two images of dimensions (64,64,3) along with a vector comprising of the gripper’s XYZ position, gripper width, and a boolean flag indicating whether the cloth is in tension between the pinned corner and the robot.

\begin{figure*}[h]
    \centering
    \includegraphics[width=\textwidth]{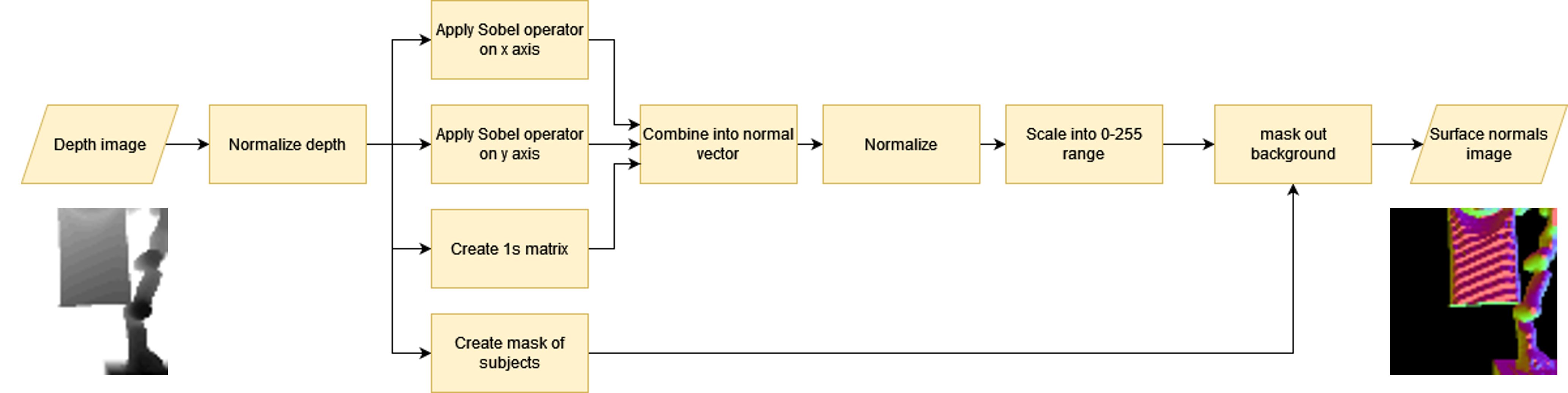}
    \caption{Pipeline of depth image to surface normals image. Higher resolution depth images output as 64x64x3 resolution normals. Same algorithm is applied in both simulation and real-world environments in real-time. The Sobel filters use a kernel size of 9 and the input/output resolution used in this process is a square 256x256 }
    \label{fig:depth to normals alg}
\end{figure*}

Reinforcement learning agents require continual feedback of reward values to determine whether the action taken was the most optimal. The reward is formulated from the surface area of the cloth and is normalized using the maximum surface area. To prevent exploitation of this reward, the Franka arm cannot pull the cloth beyond a particular threshold of strain, which is given from the environment.

\begin{equation}
\begin{aligned}
A_{ij}^{(1)} &= \frac{1}{2} \Big\| \big( v_{i,j+1} - v_{i,j} \big) \times \big( v_{i+1,j} - v_{i,j} \big) \Big\|, \quad \\[1mm]
A_{ij}^{(2)} &= \frac{1}{2} \Big\| \big( v_{i+1,j+1} - v_{i,j+1} \big) \times \big( v_{i+1,j} - v_{i,j+1} \big) \Big\|, \\[1mm]
A &= \sum_{i=1}^{3} \sum_{j=1}^{3} \Bigl( A_{ij}^{(1)} + A_{ij}^{(2)} \Bigr), \\[1mm]
A_{\text{norm}} &= \frac{A}{A_{\max}}.
\end{aligned}
\end{equation}

Formula to calculate the surface area of the cloth given a 4x4 grid of vertices. Where each vertex is given by $v_{ij} = \big( x_{ij},\, y_{ij},\, z_{ij} \big).$

\begin{equation}
R = A_{\text{norm}} - t,
\end{equation}
Final reward formula. A fully unfolded cloth (with A near $A_{max}$) will yield $A_{norm}$ close to 1, and as time progresses, the timestep penalty, t, will subtract 1 from the reward to encourage the agent to achieve an unfolded state quickly. $A_{max}$ is calculated on episode start as soon as the cloth is spawned into the environment and before it becomes effected by gravity.

\subsection{Model}
This work leverages the DreamerV2 model to teach agents how to unfold cloth objects \cite{hafner2020mastering}. DreamerV2 is a state of the art reinforcement learning model that enables agents to learn and plan in complex environments by building a so-called {\textit{world model}} -— a compact representation of the environment. Unlike traditional RL methods, which rely only on direct interaction with the environment for learning, DreamerV2 first focuses on predicting future states and rewards based on past experiences to train this world model. This approach contrasts with methods like offline reinforcement learning, where both the policy and value functions (actor-critic) are trained from pre-collected data \cite{singh2020cog} \cite{kumar2020conservative}. In DreamerV2, only the world model is initially trained, while the actor-critic components begin their learning process after interacting with the environment for the first time. This setup allows the agent to internally simulate (or ``dream'' about) possible future scenarios, enabling it to make informed decisions without the need for constant real-world interaction. The ability to plan actions within this latent space leads to more effective decision-making, especially in complex or visually rich environments. DreamerV2 has demonstrated significant success in challenging environments, including Atari games and the DeepMind Control suite, showcasing robust performance and adaptability \cite{tassa2018deepmind} \cite{yarats2021mastering}.

The Dreamerv2 model supports state inputs that combine images and vectors, allowing for multiple image views. In the context of this task, a single image view of the cloth is insufficient to capture the necessary information for effective 3D manipulation. To address this, the agent receives inputs from two cameras: a stand-off camera offering an overall view of the scene, and a wrist-mounted camera providing an egocentric perspective. The stand-off camera is crucial for assessing the cloth’s state and determining movements along the $X$ and $Z$ axes, while the wrist-mounted camera is essential for precise grasping and movements along the $Y$ and $Z$ axes.

\subsubsection{Surface normals in place of $RGB$}

Cloth objects can exhibit a wide range of shapes, sizes, colours, and textures, requiring the agent to successfully perform tasks with each variation it encounters. To address this, these attributes are randomized in each episode, along with its starting position and rotation. This variability poses a significant challenge for our world model, as the visual input of the cloth can vary almost infinitely, making it difficult for the model to compress the state into a consistent and usable latent space. For instance, a cloth object might be positioned and angled identically in two different episodes, yet differences in colour or texture could result in two distinct latent states. This discrepancy can lead the agent to take different actions in what should be similar situations, ultimately resulting in sub-optimal decisions.

\begin{figure}
    \centering
    \includegraphics[width=1\linewidth]{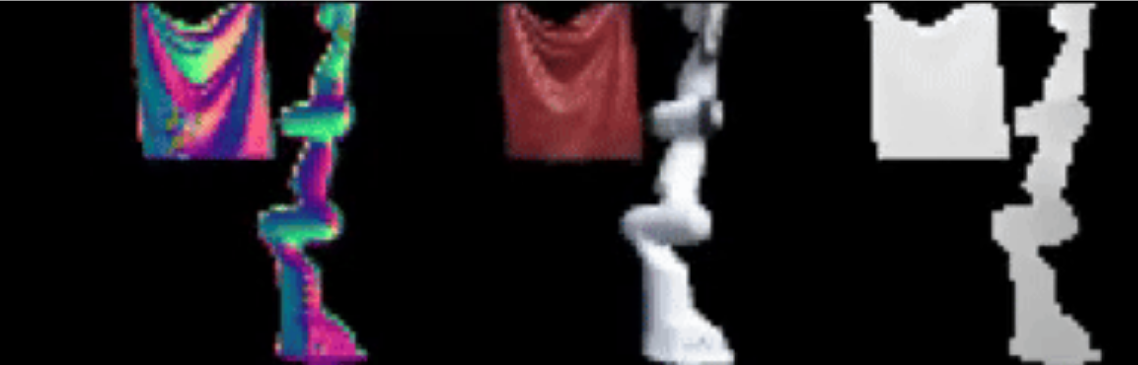}
    \caption{Simultaneous captures of the Franka robot unfolding a towel with views of surface normals, RGB, and depth, respectively}
    \label{fig:normals vs rgb}
\end{figure}

To address this issue, this work uses surface normal images rather than $RGB$. Surface normal images (or normal maps), are a type of image used primarily in computer graphics to represent the orientation of surfaces in a $3D$ space. Each pixel in a surface normal image encodes the direction that a surface is facing at that point with the $R$, $G$, and $B$ channels of the image being used to store the $X$, $Y$, and $Z$ components of the normal vector. This representation is beneficial over $RGB$ for the following reasons: the geometry of the cloth is better conveyed to the agent, and the colour/texture of the cloth is no longer a factor in the state representation. The surface normal images in this work are produced from depth images \ref{fig:depth to normals alg} and can be produced in real-time. As shown in \ref{fig:normals vs rgb}, the surface normals derived from the depth image emphasize geometric features like wrinkles and creases while disregarding the cloth's color and texture. This will help the RL agent focus on the most relevant aspects of the cloth's physical state, leading to better performance in the task of picking up and manipulating the cloth. Traditional surface-normals exhibit thin lines to represent the orientation of the surface and where light should bounce. However, on the images produces from our method, the colours blend into each other more. This is because the kernel size of the Sobel filters in our process are increased, increasing the size of the banding lines, and causing the colours to be smoother. This was done so that the images don't lose data when down-scaled to the $64 \times 64$ resolution required by the world model.

\subsubsection{Modifying the replay buffer}
Another challenge posed by our world model is image overfitting. Given the nearly infinite variations of colours and textures that can exist in the environment, it is impractical for the agent to gather training data for every possible scenario. Consequently, the agent is likely to encounter images it has never seen before, increasing the risk of overfitting. While using surface normal images might mitigate overfitting in some cases, a significant issue arises when the agent can only view images of the unfolded cloth after successfully completing the task. In the current model, an image of the unfolded cloth is required in advance to generate the corresponding latent space. In the standard Dreamerv2 model, the process begins with the agent randomly exploring the environment, using these initial episodes to train the first iteration of the world model. Only after this phase does the Dreamerv2 model start actively engaging with the environment. However, this reliance on random exploration poses a significant limitation, as it is unlikely to succeed in a complex environment like cloth manipulation and, therefore, acquire observations of the goal-state to train the world model.  

To address these overfitting issues, we modify the replay buffer. Drawing inspiration from the DRQv2 model \cite{yarats2021mastering}, which uses random augmentation to enhance the robustness of its image encoders, our approach applies random augmentations uniformly across the stack of images sampled from the replay buffer. These augmentations are applied as a batch to maintain consistency between time steps. Additionally, we fix the prevoiusly-mentioned issue with pretraining by loading a set of demonstration episodes before training begins, rather than relying on randomly explored episodes (Figure \ref{fig:unfold demo}). This ensures that the replay buffer contains images of the complete task, allowing the latent space to be accurately conveyed to the agent during high-performance episodes, thereby facilitating better exploration during training. Augmentations that may be applied to the images include adjusting the brightness/contrast or shifting the hue/saturation/values of the image or applying slight rotations, translations or zoom. We do not flip the image or overly rotate it. This ensures that the visual feedback and the axis of control are consistent. 

%% file: sections/results.tex
\section{Results}
\begin{figure*}[h]
    \centering
    \includegraphics[width=\textwidth]{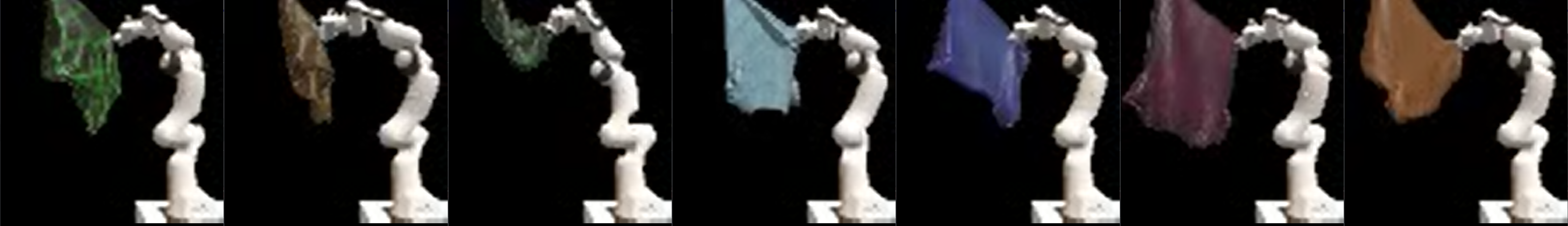}
    \caption{Example unfolds of a t-shirt garment. 2 examples of failure cases, 2 near-success cases, and 2 success cases, respectively.}
    \label{fig:success/near/fail examples}
\end{figure*}

\input{sections/results_slim}
\subsection{Simulation}
\subsubsection{Benchmarking}
\begin{figure}
    \centering
    \includegraphics[width=\linewidth]{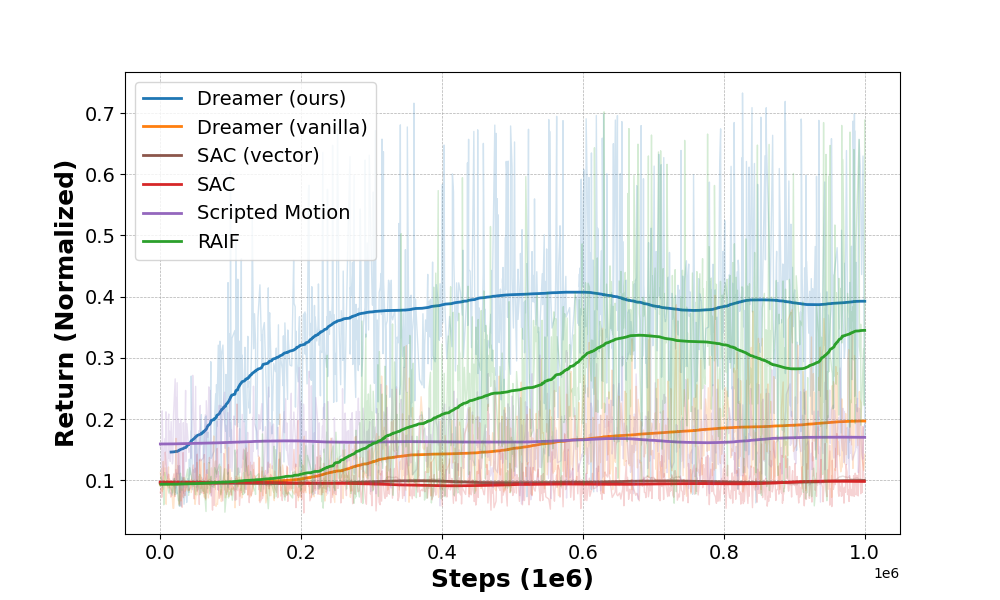}
    \caption{Smoothed chart of normalized rewards for our model, the vanilla Dreamerv2 model, an SAC, an SAC on vector inputs, and a pick/place script. All images used are Surface-Normals. Early-stopping is triggered at 80\% unfold.}
    \label{fig:normalized_reward}
\end{figure}

To demonstrate the effectiveness of our modifications, the heuristic script used to gather demonstration episodes was used as a benchmark. This gives a fair comparison against other methods in literature that utilize pick and place methods; using the known location of the corner of the cloth to pick, and the dimensions of the cloth, the script performs a pick and unfold action and has the capability to regrasp in the event that it has not been successful. Additionally, an SAC is trained with the same observation-space as our model and another is trained with vector inputs, with the XYZ positions of a select 4x4 grid of vertices of the cloth being used in place of the images. \ref{fig:normalized_reward} shows the results of the benchmarking using a normalized return. Normalization was done by dividing our episode rewards by the maximum episode timesteps and adding a 1 to remove the timestep penalty. These resulting values convey how close the cloth is to the optimal unfold state. 

While the original Dreamerv2 model manages to exceed the pick/place benchmark, our modifications provide a significant boost to the performance. The Dreamerv2 model used random exploration to achieve its performance, which does not always yield the most optimal strategy, whereas our model quickly adapted to the environment, due to the expert data loaded into the replay buffer on start, and was able to achieve a high reward very quickly and reach optimal unfold states to exploit the early stopping. We also benchmark against R-AIF \cite{nguyen2024r}, using the same demonstration episodes, achieving similar performance to our modified Dreamerv2 model. 

\subsubsection{Garment Manipulation}
We train agents according to the above-mentioned Dreamerv2 replay buffer modifications in conjunction with surface normals (in place of plain RGB input). To evaluate the effectiveness of the options, we train four different types of agent:
\begin{itemize}
    \item RGB input with vanilla replay buffer
    \item RGB input with modified replay buffer
    \item Normals image input with vanilla replay buffer
    \item Normals image input with modified replay buffer
\end{itemize}

Each agent is trained in a simulated environment for 1M timesteps. During each episode in the training loop, the cloth's position, rotation, size, color, and texture are randomized, as well as the starting position of the robot’s end effector. Besides the plane cloth, the agents are also trained on objects representing a variety of real-world garments. For each of the following garment types, a separate agent is trained for each of the four specified variations:
\begin{itemize}
    \item Plane cloth (e.g., towel, dishcloth)
    \item Short-sleeve shirt (e.g., t-shirt, buttoned-up short-sleeve shirt)
    \item Long-sleeve shirt (e.g., long-sleeve t-shirt, long-sleeve buttoned-up shirt, jumper/pullover)
    \item Shorts (e.g., cargo shorts, swim trunks, undergarments)
    \item Unbuttoned shirt (e.g., unbuttoned long-sleeve shirt)
\end{itemize}


In total, 20 agents are trained using the same parameters and evaluated at the end of 1M train timesteps. Each episode is categorized as a failure, near success, or success based on the highest measurement of unfold in the episode. The agents are also benchmarked against a pick-and-place algorithm, commonly used in the literature. This entails identifying the optimal pick point on a cloth, and executing the pick-and-place operation towards a desired configuration. In this benchmarking process, the algorithm is provided with the adjacent corner to the pinned point on the cloth as the target point. The table below shows the performance of each agent.

Table \ref{table: results-slim} presents the results of the agents performing across various garments. The highest-performing agents are those that utilize surface normal images in conjunction with the modified replay buffer. On average, these agents achieve a 15$\%$ increase in success compared to the pick-and-place algorithms and a 35$\%$ reduction in failure rates. The policies demonstrate adaptability to challenges such as cloth perturbations, difficult-to-reach corners, and incorrect state predictions. These agents also exhibit advanced behaviors such as regrasping and manipulating the cloth to expose the optimal pick points, leading to higher success rates compared to the benchmark methods, which often struggle when the cloth was not initially in a readily manipulable state. 

The main outlier in performance was observed with the shorts. None of the agents were able to match the performance of the pick-and-place benchmark. This discrepancy is likely due to the smaller size of the shorts compared to the other garments tested. Because the shorts occupy fewer pixels in the visual representation, the agents may have struggled to accurately identify optimal grasping points, leading to reduced performance.

\subsection{Real-world evaluation}
The real-world evaluation experiments are designed to closely mirror the conditions in the simulation. In these experiments, a cloth is consistently suspended by its corner in front of a Franka Emika robot, with the robot’s end-effector positioned nearly parallel to the floor. A stereo camera is suitably placed to capture a side view of both the cloth and the robot, while a second stereo camera is mounted on the robot directly before the robot’s end-effector to provide detailed imaging of the ``ego'' viewpoint. Both cameras are positioned similarly to the setting in simulation. To ensure accurate data collection, images and proprioceptive information are synchronized in real-time, followed by data formatting and conversion into the Unity coordinate system. Depth images are converted into surface normals using the same method as for the simulation experiments \ref{fig:depth to normals alg}, however occlusions in the depth images need to be patched algorithmically. The robot is controlled via a real-time cartesian impedance controller, which adjusts the end-effector's movements based on displacements provided by a pre-trained agent. This agent is loaded from a specific directory before the experiment begins. Successful unfolding attempts are recorded, and the robot is programmed to return to its initial position after a predefined number of timesteps, ensuring consistency across trials. 

Although these agents perform very well in simulation, translating the success to a real-world environment is challenging due to the depth sensors having a minimum sensing distance and the sensor noise increasing dramatically as the robot gripper, and therefore the ego-mounted camera, approaches the cloth. The surface-normals provide a rich source of information about the scene. However, a drawback imposed by the sensor range suggests that future work might benefit from another input method in place of surface-normals obtained from the ego-mounted camera. One such candidate is tactile sensing as in  \cite{sunil2023visuotactile}.
The results in table \ref{table: realtowel} show the performance of our agent trained on plane-cloth garments, deployed zero-shot into the real-world environment.

\input{sections/results_table_planecloth}

The success rates of the cloth unfolding trials are given as an average over $25$ attempts per garment. The cloth used during this experiment is swapped between trials to give a variety of colours, textures, stiffness, and sizes. The placement, pinned corner, and condition of the cloth is also adjusted between trials. For example, some trials feature a cloth with a protruding corner and others where the cloth corner is tucked away. Comparing against research with similar experimental setups, the our trials exhibit a success rate of $74$\%\. Similar expriments in the literature sometimes report success rates as high as $82$\%\ \cite{proesmans2023unfoldir} and $86.25$\%\ \cite{gabas2021dual}. However, these are not directly comparable as our trials evaluate in real-time with pre-processing required only to load our agent and not for any other stage of our trials. During real-world evaluation, the agents struggle to grasp the cloth when the corners have curled around, hence will often grasp near the corner for a near-perfect grasp. When the cloth corner was protruding, but at a tangent to the robot gripper, this would be difficult for the agent to grasp. This suggests that the biggest improvement to this work would be to apply the presented methodology onto a platform with two manipulators so that the cloth can be rotated in the air, making the corners easier for the agent to access.

Agents trained without our modifications often exhibit overfitting on static images, owing to the fixed camera positions, leading to inaccurate predictions and sub-optimal actions. This overfitting typically manifests in static elements within the images, such as the robot's base and the gripper fingers. In contrast, agents trained with our method, which applies image augmentations such as translation, zooming, and rotation to the replay buffer, do not experience these issues. Consequently, the camera placement in real-world experiments do not need to perfectly mimic the simulation, as the world model can compensate for slight offsets or changes in the field of view.

In summary, this paper presents the following key results: \begin{itemize} 
\item Expert demonstrations are essential not only for bootstrapping the learning process but also for enabling agents to acquire and incorporate effective strategies for complex cloth manipulation tasks. 
\item Replacing traditional RGB input with surface normals leads to higher-performing agents, as demonstrated across 4 out of 5 garment types evaluated. 
\item Zero-shot deployment of a trained agent is feasible, enabling real-time unfolding of cloth objects without additional training. 
\end{itemize}

%% file: sections/results_slim.tex
\begin{table*}[]
\centering
\resizebox{\textwidth}{!}{%
\begin{tabular}{@{}llllllllllllllllllllllll@{}}
\toprule
\textbf{} &
  \multicolumn{3}{c}{\textbf{Long-sleeve}} &
   &
  \multicolumn{3}{c}{Short-sleeve} &
   &
  \multicolumn{3}{c}{Plane-cloth} &
   &
  \multicolumn{3}{c}{Shorts} &
   &
  \multicolumn{3}{c}{Unbuttoned shirt} \\ \midrule
\textbf{RGB vanilla} &
  45.6 &
  20.4 &
  \textbf{34.0} &
   &
  33.2 &
  24.4 &
  42.4 &
   &
  64.8 &
  33.2 &
  2.0 &
   &
  38.4 &
  51.2 &
  10.4 &
   &
  64.8 &
  32.0 &
  3.2 \\
\textbf{RGB (ours)} &
  40.0 &
  13.6 &
  46.4 &
   &
  31.6 &
  19.2 &
  49.2 &
   &
  34.4 &
  46.0 &
  19.6 &
   &
  44.0 &
  32.8 &
  23.2 &
   &
  33.3 &
  33.3 &
  33.3 \\
\textbf{normals vanilla} &
  82.4 &
  12.8 &
  4.8 &
   &
  17.2 &
  19.2 &
  63.6 &
   &
  12.0 &
  52.0 &
  36.0 &
   &
  22.4 &
  44.0 &
  33.6 &
   &
  60.8 &
  29.6 &
  9.6 \\
\textbf{normals (ours)} &
  \textbf{4.0} &
  \textbf{24.0} &
  \textbf{72.0} &
   &
  \textbf{0.0} &
  \textbf{0.0} &
  \textbf{100} &
   &
  \textbf{0.0} &
  \textbf{15.2} &
  \textbf{84.8} &
   &
  44.8 &
  37.6 &
  17.6 &
   &
  \textbf{29.6} &
  \textbf{28.0} &
  \textbf{42.4} \\
\textbf{pick/place} &
  35.8 &
  0.0 &
  64.2 &
   &
  36.0 &
  0.0 &
  64.0 &
   &
  71.2 &
  0.0 &
  28.8 &
   &
  \textbf{44.8} &
  \textbf{0.0} &
  \textbf{55.2} &
   &
  96.8 &
  0.0 &
  3.2 \\ \bottomrule
\end{tabular}
}
\caption{Table of evaluation results for garment unfolding. Percentage result Fail/Near/Success for 250 trials per agent. }
\label{table: results-slim}
\end{table*}

%% file: sections/results_table_planecloth.tex

\begin{table}[]
\centering
\begin{tabular}{@{}lcccll@{}}
\toprule
\textbf{}           & \multicolumn{1}{l}{\textbf{Failure}} & \multicolumn{1}{l}{\textbf{Success}} & \multicolumn{1}{l}{\textbf{Avg Timesteps}} &  &  \\ \midrule
\textbf{towel}      & 28\%                                 & 72\%                                 & \textbf{107}                               &  &  \\
\textbf{dishcloth}  & 32\%                                 & 68\%                                 & 209                                        &  &  \\
\textbf{facecloth}  & 24\%                                 & 76\%                                 & 212                                        &  &  \\
\textbf{pillowcase} & 20\%                                 & \textbf{80\%}                        & 162                                        &  &  \\
\textbf{Average}    & 26\%                                 & 74\%                                 & 172                                        &  &  \\ \bottomrule
\end{tabular}
\caption{Results of real-world deployment of planecloth unfolding agent. Percentage success averaged over 25 trials per garment - failure is marked if an unfold is not completed within the designated timesteps or if the agent performs a failed unfold and does not release the cloth}
\label{table: realtowel}
\end{table}

%% file: sections/conclusion.tex
\section{Conclusions}
This work demonstrates the benefits of scaling up world models within a modern reinforcement learning (RL) framework to address the challenging robotic task of in-air cloth unfolding. A key component of this scaling is the use of surface normals, computed in real-time from depth data, in place of traditional RGB inputs. This input modality, combined with several architectural and training modifications, led to more robust and higher-performing agents.

In particular, augmenting the replay buffer with demonstration episodes, which are used in place of initial exploratory episodes, and applying randomized image augmentations significantly improved performance on the unfolding task. We benchmark against similar works (R-AIF) as well as the vanilla Dreamerv2, a pick/place script, and standard SAC models. We show that despite being comparable to the newer Dreamerv3 \cite{hafner2023mastering}, we can achieve similar (if not slightly better) performance to R-AIF using the Dreamerv2 model with our modifications. We also prove with each model, the effectiveness of using RL agents for control strategy as opposed to traditional pick/place approaches.

Our models were trained and evaluated within a custom Unity simulation environment featuring realistic cloth physics and high-fidelity rendering. Importantly, the trained agents achieved comparable performance to prior work when deployed on real robot hardware—while operating in real time, without the need for extensive pre- or post-processing. These results further highlight the potential of RL for complex, real-world manipulation tasks, particularly in domestic and service robotics domains.

Across all garment types, our system achieved an average unfolding success rate of 74\%. While promising, this level of performance was likely impacted by a high degree of noise in the egocentric depth data. In particular, when the robot arm approached the cloth, the depth camera often came closer than the sensor's minimum range, resulting in incomplete or heavily distorted observations. These out-of-distribution inputs significantly hindered the model during deployment, especially during critical manipulation phases. Future work should explore improved sensor placement or alternative camera configurations to ensure consistent visual input throughout the task. The ego-centric camera worked best during trials, compared to other second camera placements, however, a different strategy is required to collect ego-centric depth data for surface normals. Additionally, RGB-to-depth models were also trialed, but had similar results to raw depth. 

Other limitations also emerged. The agent underperformed on the shorts garment, likely due to the garment's small size and limited pixel representation within the 64×64×3 observation space. This reduced spatial footprint impairs both perception and action planning. Solutions such as higher-resolution input, spatial attention mechanisms, or adaptive cropping could help mitigate this challenge.

Additionally, due to the constraints of our single-arm Franka platform, we did not include large garments (e.g., trousers, dresses, or plus-sized clothing) in our evaluation. We hypothesize that applying our methods to a dual-arm robot (e.g., Baxter or UR series) would enable successful manipulation of such garments, particularly by enabling mid-air repositioning and coordinated gripping strategies.

Finally, each garment type required a separately trained agent for optimal performance. While effective, this approach is both time-intensive and resource-intensive. Future work should focus on policy transfer and generalization; developing methods that allow for shared representations and efficient training across a broader range of garment types.
